 \documentclass[sigconf,nonacm]{acmart}
\usepackage{floatpag}
\floatpagestyle{empty} 

\usepackage{color}

\newcommand{\IR}{LLVM~IR}

\copyrightyear{2023}
\acmYear{2023}
\setcopyright{rightsretained}
\acmConference[GECCO '23 Companion]{Genetic and Evolutionary Computation Conference Companion}{July 15--19, 2023}{Lisbon}
\acmBooktitle{Genetic and Evolutionary Computation Conference Companion (GECCO '23 Companion), July 15--19, 2023, Lisbon, Portugal}\acmDOI{10.1145/3583133.3590542}
\acmISBN{979-8-4007-0120-7/23/07}

\newlength{\halfwidth}
\begin{document}

\title{GI Software with fewer Data Cache Misses} 
\titlenote{Long version of
W.B. Langdon, Justyna Petke, Aymeric Blot, David Clark. 2023.
Genetically Improved Software with fewer Data Cache Misses.
In Genetic and Evolutionary Computation Conference Companion
(GECCO ’23 Companion), July 15–19, 2023, Lisbon, ACM.
\url{https://doi.org/10.1145/3583133.3590542}
}

\author{\href{http://www.cs.ucl.ac.uk/staff/W.Langdon}
{W.B. Langdon},
\href{http://www.cs.ucl.ac.uk/staff/J.Petke/}
{Justyna Petke},         
\href{https://www-lisic.univ-littoral.fr/author/blot/}
{Aymeric Blot},          
\href{http://www.cs.ucl.ac.uk/staff/D.Clark}
{David Clark}            
}
\email{{W.Langdon,j.petke,david.clark}@ucl.ac.uk}
\email{aymeric.blot@univ-littoral.fr}

\affiliation{%
  \institution{Department of Computer Science, University College London}
  \streetaddress{Gower Street, London, UK}
  \city{}
  \state{}
  \country{}
  \postcode{WC1E 6BT, UK}
}

\begin{abstract}
By their very name caches are often overlooked and yet play a vital role
in the performance of modern and indeed future hardware. 
Using
\href{https://github.com/bloa/magpie}
{MAGPIE}
(Machine Automated General Performance Improvement via Evolution of software)
we show genetic improvement~GI can 
reduce the cache load of existing computer programs.
Operating on lines of C and C++ source code
using local search,
Magpie can generate new functionally equivalent variants
which generate fewer L1 
data cache misses.
Cache miss reduction is tested on 
two industrial open source programs
(Google's Open Location Code~%
\href{https://github.com/google/open-location-code}{OLC}
and Uber's 
Hexagonal Hierarchical Spatial Index~%
\href{https://github.com/uber/h3}{H3})
and two 2D photograph image processing tasks,
counting pixels
and OpenCV's SEEDS segmentation algorithm.

Magpie's patches functionally generalise.
In one case they 
reduce data misses on the highest performance L1 cache dramatically by~47\%.
\end{abstract}



\maketitle

\section{Introduction}

Jack Dongarra won the Turing Award in 2021.
In celebration,
in a recent article in the Communications of the ACM~%
\cite{DBLP:journals/cacm/Dongarra22}
he describes the current and foreseeable future limits of
high performance computing considering that clock frequencies have not
increased significantly for two decades but that Moore's Law~%
\cite{Moo65}
continues to provide exponential increases in transistor count.
Dongarra argues that this will continue to fuel the
current trend to ever greater degrees of hardware parallelism.
He also points out that already computing hardware is limited
not by processor speed but by the time taken to get data to
the compute engines.
That is, computing is cheap, it is data movement that is expensive.

Although widespread,
the trend to evermore hardware parallelism
is exemplified by graphics cards (GPUs) and
similar accelerator architectures (e.g.~TPUs)
which now form the back bone of both super computers
and training deep neural networks.
For a long time nVidia were reluctant to introduce general data caches
into their graphics cards,
instead insisting the developers of computer games would
know the data flows inside their programs and could therefore
optimise their software to take best advantage of the huge data
bandwidth available in GPUs.
This was always unrealistic and,
even for very experienced programmers,
getting the best performance was very hard.
Hence nowadays
in an attempt to make performance programming easier,
GPU manufactures
boast that their GPUs contain multiple levels of cache memory.

Dongarra tries to argue against the trend to use caches
to make life easier for software developers,
and says
``Instead of just relying on hardware caches, new algorithms must be
designed''~%
\cite[page~68]{DBLP:journals/cacm/Dongarra22}.
However, he does not say how this will be done.
Genetic programming has a long established tradition
of inventing new solutions~\cite{koza:1999:GPdim}.
In contrast to GP,
rather than starting evolution from scratch
every time,
Genetic Improvement (GI)~\cite{Petke:gisurvey,langdon:2010:cigpu}
finds updates on existing software.
Although primarily used for fixing computer bugs~\cite{DBLP:conf/gecco/ForrestNWG09}
or speeding up code~\cite{Langdon:2013:ieeeTEC},
we seek to strengthen the claim that
it can optimise any {\em measurable} aspect
of software by
showing that, despite noise,
the GI tool
Magpie 
can sometimes increase the effectiveness of even the fastest
level of data caches in a conventional Intel x86 desk top computer,
even on compiler optimised code%
\footnote{%
CPU hardware provides both data and instruction caches.
These are both fundamental to performance.
However typically good locality and the small size of program machine code
compared to data it acts upon,
means instruction caches now
and probably into the future,
are less of an issue than data caches.
In these experiments with an L1~instruction cache of 32KBytes
there are only about a hundred instruction cache misses,
while, for example,
with SEEDS the number of L1 data misses exceeds 40\,000
even on a small image
such as Figure~\ref{fig:seeds_training}.
}.
Thus GI and GP may form a two pronged attack on the problem of
effectively using future data paths in highly parallel hardware,
with perhaps GP inventing new algorithms
and GI tailoring existing code to take better advantage of novel hardware.

\vspace{1ex}
Section~\ref{sec:fitness} details the fitness test cases
for Google's OLC and Uber's H3 digital mapping programs,
the Blue image benchmark,
and the OpenCV SEEDS resource intensive image segmentation algorithm.
While Section~\ref{sec:search} describes 
using a Coupon Collector argument to choose
how much of the search space Magpie should sample.
Following the results in Section~\ref{sec:results}
(see also Table~\ref{tab:cache_tab}),
in Section~\ref{sec:discuss} there is
a brief description of the C and C++ source code changes made by Magpie,
which in the image examples
give a reduction in L1 data cache misses
even on the existing compiler (GCC~-O3) optimised code,
and a discussion of non-stationary noise. 
We conclude
in Section~\ref{sec:conclude}
that Magpie is ready to use
and here it reduced L1 data cache misses by up to 47\%.
In all four examples the patches functionally generalise
but only in the data hungry image processing examples
do we see any sustained cache reduction
(see Figures~\ref{fig:uk},
\ref{fig:seeds_holdout},
\ref{fig:validate_olc_l1},
\ref{fig:validate_h3_l1},
\ref{fig:blue_l1_patch} and
\ref{fig:seeds_l1_O3_patch}).
But first we describe the background.

\section{Background}

\noindent
Computers are universal.
Everything relies on them.
Even now many of the richest people
are rich because they
founded very successful software companies
at about the turn of the century. \linebreak[3] 
IT technology in general and software in particular
permeates and will continue to dominate
the third millennium.
Indeed the planet is addicted to software.
Despite programming being more than 60 years old,
software continues to be hand written.
Automatic programming to a large extent remains a dream.

At GECCO-2009 Stephanie Forrest,
ThanhVu Nguyen, Wes Weimer and Claire {Le~Goues} 
\cite{DBLP:conf/gecco/ForrestNWG09}
\nocite{Weimer:2009:ICES}
showed that genetic programming 
\cite{koza:book,banzhaf:1997:book,poli08:fieldguide,%
ryan:book,%
book:1999:aigp3,%
Vanneschi:book}
could automatically fix bugs
in computer software~%
\cite{Weimer:2010:ACM,cacm2019_program_repair}.
For the first time artificial intelligence (AI) is being applied
to a major problem in software engineering on industry size
programs~\cite{Alshahwan:2019:GI}.
Since then the field of automatic program repair~(APR) has bloomed~%
\cite{Monperrus:2018:ACM_CSUR}.
Inspired by Stephanie et~al.\ 
we~\cite{langdon:2010:cigpu,Langdon:2013:ieeeTEC}
began applying genetic programming to improving human written
software in many ways in addition to bug fixing~\cite{Petke:gisurvey}.

Although genetic programming remains a common
search technique in genetic improvement (GI),
local search is increasingly popular~\cite{blot:2021:tevc}.
In addition to ever more powerful computers,
GI is able to scale because it 
does not start from scratch at every run but
builds on existing software.
Another enormous advantage is that
GI can automatically double check with
the existing painstakingly hand written code.
As with regression testing~\cite{Harman:2013:WCRE},
in effect, the existing program 
becomes its own specification.
That is, it can be used as a test oracle for 
both automatic functional and non-functional improvements.
This could be run time performance
(be it elapse time, memory requirements, etc.)
and also
its functionality.
We have mentioned above the success of automatic bug fixing,
but functional improvements can include
making the program give more accurate answers~%
\cite{langdon:2018:EuroGP}.
Evolution has also been used to adapt web software to
colour blind users,
and to help tune hearing aids for deaf patients~\cite{Legrand:2007:GPEM}.
The increasing success of automatically generated software tests~%
\cite{fraser:evosuite}
also naturally feeds into genetic improvement.

Genetic improvement
research is often via bespoke one-off experiments.
David White recognised this and proposed
GIN~\cite{White:2017:GI} as a generic GI tool for Java programs.
Gabin An~\cite{An2017aa,An:2018:GI} proposed PyGGI for Python.
Both tools have been extensively used and updated:
\cite{%
Brownlee:2019:GECCO,%
Petke:2019:SSBSE,%
Petke:2020:APR,
DBLP:journals/corr/abs-2202-01490}
and
\cite{%
an:2019:fse,%
Kitt:2021:GI,%
Smigielska:2021:GI,%
Mesecan:2021:GI,%
blot:2021:tevc%
},
and further GI tools have been proposed \cite{Krauss:2022:GI}. 
Nevertheless recently
a user study said that GI lacked user friendly tools~%
\cite{Zuo:2022:GI}.

\subsection{Magpie}

In response to this Aymeric Blot wrote Magpie, 
which is not only user friendly but combines GI and parameter tuning.
It was released last year
as an open source project.
Like PyGGI~2.0~\cite{an:2019:fse}
(from which it was developed)
it is freely available from GitHub%
\footnote{\url{https://github.com/bloa/magpie}}.
We use it to hopefully convince the reader that evolution
can in principle improve any {\em comparable} measure of software quality.
Although also written in Python,
it aims to work with any computer programming language.
It has been mostly tested on Apple Mac and Linux Laptops
but aims to be generic enough to work under
Microsoft windows.
Here we test it on a Linux desktop
and have not attempted to maintain compatibility with Microsoft.

As of 27 November 2022,
including examples and documentation, Magpie contains
4871 lines of code, mostly written in Python.
It contains examples in Python, C, C++ and Ruby.

\subsection{Updates to Magpie}
\label{sec:q1_diff}

In the course of previous work
\cite{langdon:2023:GI},
we had enhanced Magpie
to use Python's {\tt ctypes} to directly call the patched code.
This allows us to reduce noise by
collecting data directly on the individual C/C++ routines
rather than the whole Magpie sub-process.
Also to exclude the Python interpreter from our measurement,
we clear the L1 data cache before invoking the patch code
by setting and reading a fixed array of 32K~bytes
(the size of our CPU's L1 data cache).
To make the fitness robust,
each patch is tested multiple times 
and since the mean is notoriously suspect to noisy outliers,
we use instead the first quartile to summarise the measurements.
\label{p.Q1}

In the image segmentation example (Section~\ref{sec:OpenCV})
17\% of the lines consist of a single closing brace~\verb'}'.
These are naturally interchangeable and so it
was discovered that Magpie was wasting a lot of effort testing
programs that were identical (apart from white space, etc).
To prevent this, 
the compilation step keeps a ``tabu'' list~\cite{Langdon:2015:GECCO}.
Previously~\cite{Langdon:2015:GECCO} we had
a complicated tabu of both genotypic and phenotypic information
(of up to 340~MBytes),
here we simply keep a copy of each object file
(on average 407 files per run,
occupying about 30~MBytes).
After compilation, semantically identical patches
are rejected by simply comparing their object file
with object files of the same size from previous patches.
New unique patches are added to the tabu directory and
identical ones are discarded without fitness testing.
Although in the limit this could be slow as the tabu directory grows,
in practice the time taken is negligible.

The Linux GNU perf utility allows access to many 
hardware performance counters.
In particular
we used the perf run time library 
{\tt linux/perf\_event.h} 
to collect L1 data and instruction cache misses,
the count
of instructions executed and elapsed time.
(Only the L1 cache data misses are used by the fitness function.)

\subsection{Datasets Background}

\begin{figure} 
\centerline{\includegraphics{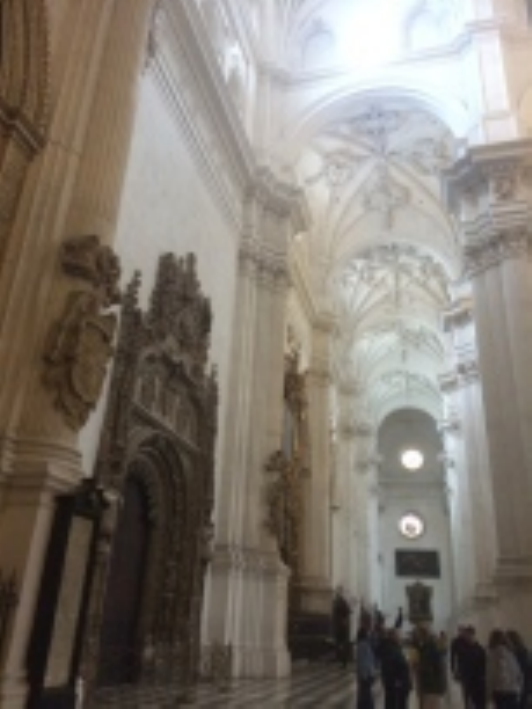}}
\vspace*{-2ex}
\caption{\label{fig:seeds_holdout}
Example holdout 2448$\times$3264 image
(tested at 100\% 
$\frac{1}{2}$,
$\frac{1}{4}$,
$\frac{1}{8}$ and $\frac{1}{16}$ sizes).
One of 30 
SEEDS examples randomly selected from
\url{http://www.cs.ucl.ac.uk/staff/W.Langdon/egp2014/granada/}
The automatically evolved SEEDS C++ code was
validate multiple times at five resolutions on each.
}
\end{figure}

We use four open source C/C++ examples.
Two industrial geospatial programs, both written in C
\cite{langdon:2023:EuroGP}.
One from Google's OLC and the other from Uber's~H3
(see Section~\ref{sec:postcodes} and Figure~\ref{fig:uk}).
And
two C++ photographic image processing examples
(Section~\ref{sec:images}).
These are:
our Blue benchmark
from the 2018 Tarot summer school~\cite{langdon:RN1806}
(Figure~\ref{fig:blue_training})
and OpenCV's image segmentation code~\cite{Langdon:2016:SSBSE}
(Figures~\ref{fig:seeds_training} and~\ref{fig:seeds_holdout}).

\section{Fitness Function}
\label{sec:fitness}
\noindent
Magpie attempts to run the patched program
on all the test cases.
If the patch passes them all,
Magpie runs it again multiple times to try to get a good robust
estimate of its performance.

In summary:
Magpie uses multiple objectives to calculate a mutation's fitness.
In sequential (priority) order:
1)~does the patch compile without error
(warnings are ignored),
2)~does the mutant software run without crashing or timing out
on every test case,
3)~are its outputs the same as those of the original code
and
4)~how many L1 data cache misses does {\tt perf} record.

(1)~The source code is compiled using the GCC compiler (version~10.2.1)
with the same options and switches, e.g.~-O3, as the developers 
(Google, Uber, the OpenCV team) use.
To avoid wasting time on reporting multiple errors,
{\tt -fmax-errors=1} is used to stop GCC on the first error.
If the compiler succeeds in compiling the patch,
it is linked with non-evolvable code outside the patch,
and the C~code that calls the {\tt perf} run time library,
to form a shared object library.
(The GCC options {\tt -shared} and {\tt -fPIC}
are used to create the shared library {\tt prog.so}.)
The Python interpreter 
uses python {\tt ctypes} on the shared library
to call the C interface routine
which calls the {\tt perf} runtime library and 
the patched code.
After the patch has been run,
the {\tt perf} runtime library
extracts hardware counters from within the CPU,
which the interface code passes back to the Python interpreter,
along with the outputs generated by the patch.

(2)~Both Magpie (via the Python interpreter)
and the mutant itself, can signal a problem via the Unix
exit status.
In either case, the main Magpie thread will 
discard the patch 
giving it poor fitness
and then move onto generating and testing the next patch.

(3)~For each test case 
the Python subprocess will check that output of the patch
is as expected
for each user supplied test case.
For example, with OLC,
Magpie checks that the patched code returned the same 16 characters
as Google's code for the test's pair of latitude and longitude.
If any characters are different or missing the test fails
and fitness testing for that patch stops immediately.
Before using
Magpie, we ran the original OLC, H3, Blue and SEEDS programs
on each test case, recorded their output
and then this was automatically converted into 
a Python list data structure.
For example, with OLC,
the fitness training consists of ten latitude and longitude floating
point numbers and ten 16~character strings,
formatted as ten Python bracketed tuples.

(4)~Section~\ref{sec:q1_diff} above describes how
the {\tt perf} C runtime library is integrated into Python.
Magpie 
uses the first quartile (Q1) of all the patch's repeated measurements
to give its fitness measure
(see also Section~\ref{sec:q1_diff}). 
Even in supposedly deterministic programs,
the hardware counters for cache statistics,
instructions run
and elapsed time are noisy.
Despite the use of robust statistics like Q1 on 
{\tt perf}s L1 data cache misses,
fitness remains noisy.
But as we will see,
in some cases Magpie is able to make progress.

\subsection{Test Cases for Google's OLC and Uber's H3: GB Post Codes}
\label{sec:postcodes}

We used the same test cases as before~\cite{langdon:2023:EuroGP}
when optimising OLC and H3.

Both
Google's Open Location Code~(OLC)
\url{https://github.com/google/open-location-code}
(downloaded 4 August 2022)
and
Uber's Hexagonal Hierarchical Geospatial Indexing System~(H3)
\url{https://github.com/uber/h3}
(downloaded the previous day) 
are open industry standards
(total sizes
OLC 14\,024
and
H3~15\,015
lines of source code).
They include C programs which convert
latitude and longitude
into their own internal codes 
(see Table~\ref{tab:cache_tab}). 
For OLC we used Google's 16  character coding
and for H3 we used Uber's highest resolution (\mbox{\tt -r 15})
which uses 15 characters.
Following our earlier work~\cite{langdon:2023:EuroGP},
we use as test cases the position of actual addresses.

\begin{figure} 
\hspace*{-1.25em}
\centerline{\includegraphics{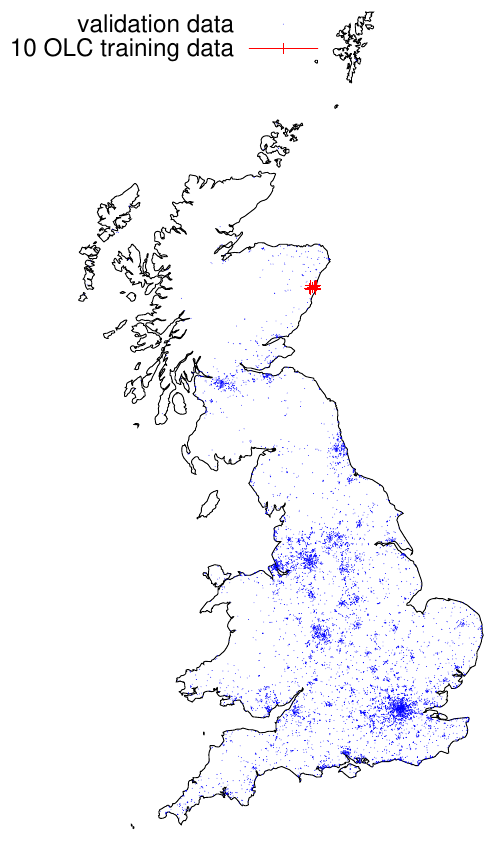} \hspace*{-1.25em} \includegraphics{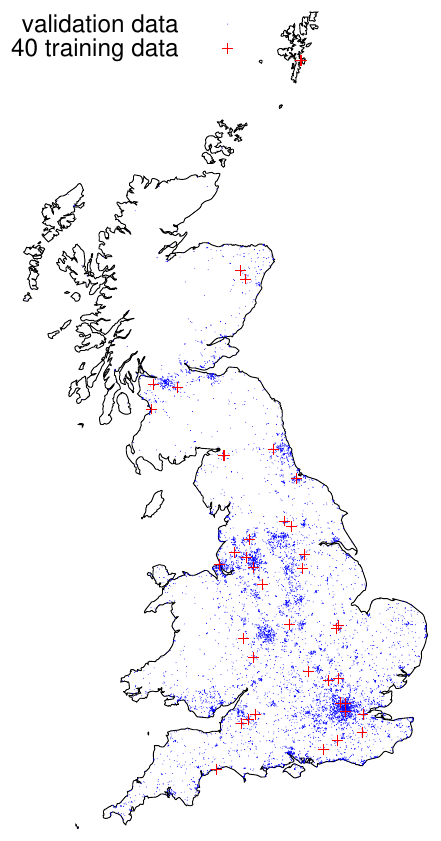}} 
\vspace*{-2ex}
\caption{\label{fig:uk}
Left:
Ten OLC training points randomly selected in the neighbourhood
of Aberdeen~(red).
Holdout set~(blue dots)
GB post codes.
Right:
Forty training points randomly selected from ten H3 runtime classes.
Holdout set~(blue dots),
locations of ten thousand random GB post codes
(no overlap with H3 training or OLC (left) holdout data).
OLC and H3 
patches pass all 10\,000 holdout tests.
Dataset from \protect\cite{langdon:2023:EuroGP}.
}
\end{figure}

For Google's OLC 
we used the~\cite{langdon:2023:EuroGP} dataset which was 
the location of the first ten thousand GB postcodes 
downloaded from
\url{https://www.getthedata.com/downloads/open_postcode_geo.csv.zip}
(dated 16 March 2022).
The addresses are alphabetically sorted starting with AB1~0AA,
which is in Aberdeen.
For training 
ten pairs of latitude and longitude were selected
uniformly at randomly
(see Figure~\ref{fig:uk}).
The unmutated code was run on each pair and its output saved
(16 bytes).
For each test case each mutant's output is compared with 
the original output.

Uber's
H3 was treated similarly
(see right of Figure~\ref{fig:uk})

\subsection{Test Cases for Blue and OpenCV SEEDS}
\label{sec:images}

\begin{figure}
\centerline{\includegraphics{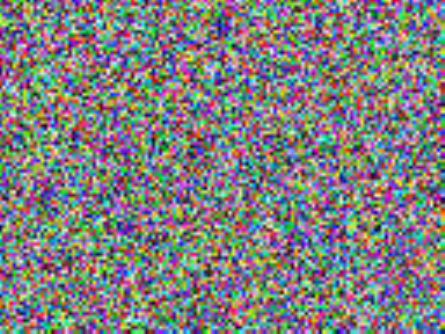}}
\vspace*{-2ex}
\caption{\label{fig:blue_training}
One of ten ``Blue'' random 96$\times$128 images.
Example contains 2135 blue pixels
\protect\cite{langdon:RN1806}.
}
\end{figure}

We had previously produced a simple example of the GISMO
GI system for students attending the 
\href{https://wp.cs.ucl.ac.uk/tarot2018/}
{2018 TAROT summer school}
on Software Testing, Verification and Validation
\cite{langdon:RN1806} 
\url{http://www.cs.ucl.ac.uk/staff/W.Langdon/ftp/gp-code/opencv_gp.tar.gz}.
We generated ten random images
(see Figure~\ref{fig:blue_training})
and calculated the number of ``blue'' pixels in each.
These were used by Magpie as training data,
with a goal of optimising their code to minimise L1 cache misses.
Notice the training images contain
$96\times128$ = 12\,288 coloured  pixels,
occupying 49\,152 bytes, and so exceed the L1 data cache.
We removed the comments,
leaving 100 lines of C++ code.

\label{sec:OpenCV}
In contrast OpenCV is an enormous suite of C++ image processing tools.
At the beginning of 2023 
OpenCV's open source repository on 
\href{https://github.com/opencv/opencv}{GitHub}
comprised
more than two million lines of code 
(mostly C, C++ and XML).
Therefore, we selected an important routine:
the state-of-the-art OpenCV SEEDS superPixels image segmentation algorithm. 
This figured in the 
\$50K OpenCV Challenge,
and we had previously
used it in GI experiments to reduce run time
whilst respecting its API~%
\cite{Langdon:2016:SSBSE}.

\begin{figure}
\centerline{\includegraphics{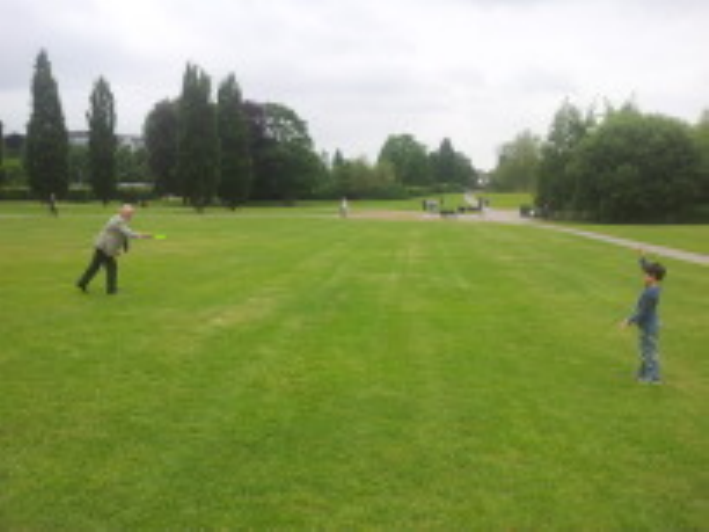}}
\vspace*{-2ex}
\caption{\label{fig:seeds_training}
OpenCV SEEDS image segmentation 
training 204$\times$153 image.
Downloaded from 
\url{http://www.cs.ucl.ac.uk/staff/W.Langdon/ftp/gp-code/opencv_gp.tar.gz}
\protect\cite{Langdon:2016:SSBSE}.
}
\end{figure}

The OpenCV code of the SEEDS SuperPixel algorithm is
1500 lines of C++ code, 
but the important routines are held in one file
{\tt updatePixels.cpp}.
After removing comments and empty lines
there are 319 lines.
Unfortunately the SEEDS algorithm is compute intensive
and so instead of using full images
obtained from~\cite{Langdon:2016:SSBSE},
we reduced the training data to $1/16$
(see Figure~\ref{fig:seeds_training}).
Notice the 204$\times$153 training image
contains 31\,212 coloured pixels
(124\,848 bytes)
and so the major data structure used by the SEEDS algorithm
exceeds the L1 data cache.

\section{Magpie Search}
\label{sec:search}

Magpie allows a clean separation of
user supplied parameters for each experiment
from its generic code.

It is often the case that there is a ``settling~in'' period
when programs take longer to run than usual.
Magpie solves this by starting with a ``warm~up''.
Before starting any search Magpie generates empty patches
and tests them.
(Originally Magpie created 4 empty patches,
we now use 11.)
The fitness of the warm up patches are discarded, 
except that Magpie reports the performance of all later patches
as a proportion of the average warm-up fitness.
Note therefore that lower Magpie fitness are better.

In addition to the 11 warm-up patches,
Magpie generated 700~OLC,
19\,077~H3,
904~Blue and
3151~SEEDS patches
(see Table~\ref{tab:cache_tab}).
The OLC and H3 (700 and 19077) values are taken from 
our previous work~\cite{langdon:2023:EuroGP}.
As before~\cite{langdon:2023:EuroGP},
we use a coupon collector~\cite{feller:1957:ipta}
argument to calculate how many random samples
would be needed to be almost certain of visiting every line
of the C/C++ source code at least once.
(The H3 source code to be optimised is much bigger than the others,
see Table~\ref{tab:cache_tab},
hence the larger search effort.)
In all four cases we used Magpie's run time reduction option:
\mbox{\tt python3 -m bin.magpie\_runtime}.

\begin{figure}
\centerline{\includegraphics{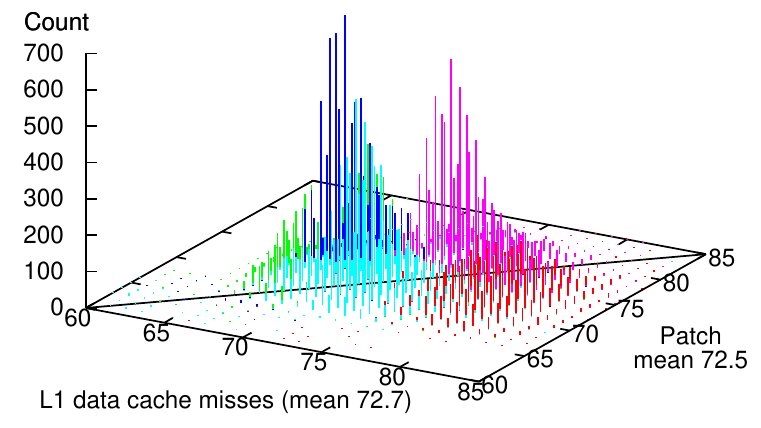}} 
\vspace*{-2ex}
\caption{
Magpie OLC patch sample performance
on 10\,000 hold out postal addresses.
The patch passes all 10\,000 examples.
Unlike H3, OLC is quite happy to process without error
the 85 invalid locations.
The 10\,000 tests have been repeated five times
on the same computer.
The presence of five distinct distributions is shown
by five colours.
In each repeat,
the original and patch have significantly different distributions,
But this is an {\em artifact},
see Section~\protect\ref{sec:discuss2}.
Left hand side of Figure~\protect\ref{fig:uk}
shows the 10 training data location and these 10\,000 holdout post codes.
\label{fig:validate_olc_l1}
}
\end{figure}

\begin{figure}
\centerline{\includegraphics{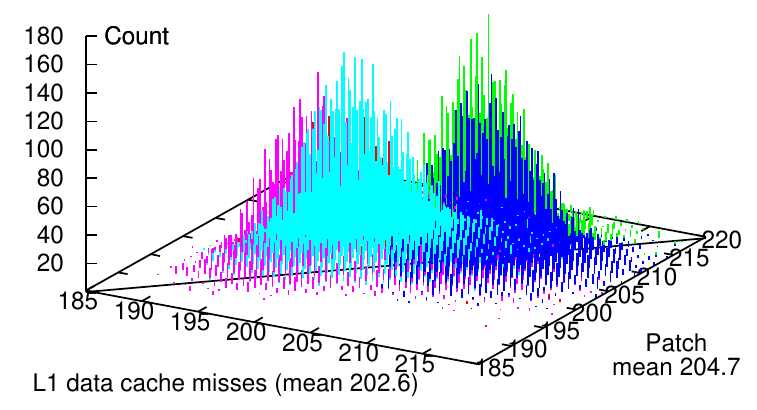}} 
\vspace*{-2ex}
\caption{
Magpie H3 patch out of sample performance
on 9915 hold out postal addresses.
The patch passes all 10\,000 examples.
There are 85 invalid locations
(for which the patch correctly generates an error message)
which are excluded from this comparison.
The patch incurs on average 2.1 extra L1 data cache misses.
Note spread of data $\sigma$=5. 
As with Figure~\protect\ref{fig:validate_olc_l1}
we use five colours to show five distinct distributions 
cause by repeating the 9915 tests five times.
The right hand side of Figure~\protect\ref{fig:uk}
shows the 40 training data location and these 9915 holdout post codes.
\label{fig:validate_h3_l1}
}
\end{figure}

Magpie used a single main thread
on an otherwise mostly idle
32~GB
3.60~GHz Intel i7-4790 desktop CPU
running networked Unix Centos~7,
using Python~3 
version~3.10.1
and
version 10.2.1 of the GNU C compiler.
In all four cases there was a lot of variation
in values recorded by {\tt perf}.
(For example, see 
Figures~\ref{fig:validate_olc_l1} and~\ref{fig:validate_h3_l1}.)

\begin{table} 
\caption{
\label{tab:cache_tab}
Left:
size of C/C++ sources 
to be optimised
(comments and blank lines removed).
Middle:
averages for up to five Magpie runs.
Columns~3--4 size of patch.
Column~5 best fitness
(average reduction in L1 data cache misses).
Right:~%
Column~6 size of search space explored.
Column~7 fraction of mutants which compile, run ok and give correct answer.
Column~8
average Magpie run times for 1 core on a 3.6~GHz Intel i7-4790 desktop
with GCC~10.2.1
and Python~3.10.1
}
\vspace*{-2ex}


\centerline{%
\begin{tabular}{@{}lr|rr|rrr@{}}
Example &LOC &
\multicolumn{2}{|c}{Mutant} &
\multicolumn{3}{|c}{Magpie}
\\
    &     &
size & L1D & 
steps & \% run ok & duration \\
\hline
OLC  & 134 & 1-- 5 &  5\% &
700   & 23\% &66 secs
\\
H3   &1615 & 6--18 &  6\% &
19077 & 33\% & 3.9 hours
\\
blue & 100 & 7--10 & 47\% &
904   & 30\% & 1.7 hours
\\
SEEDS& 319 & 2-- 7 &  7\% &
3151  &  2\% & 5.2 hours
\\
\end{tabular}
}

\end{table}

\section{Results}
\label{sec:results}

\begin{figure}
\centerline{\includegraphics{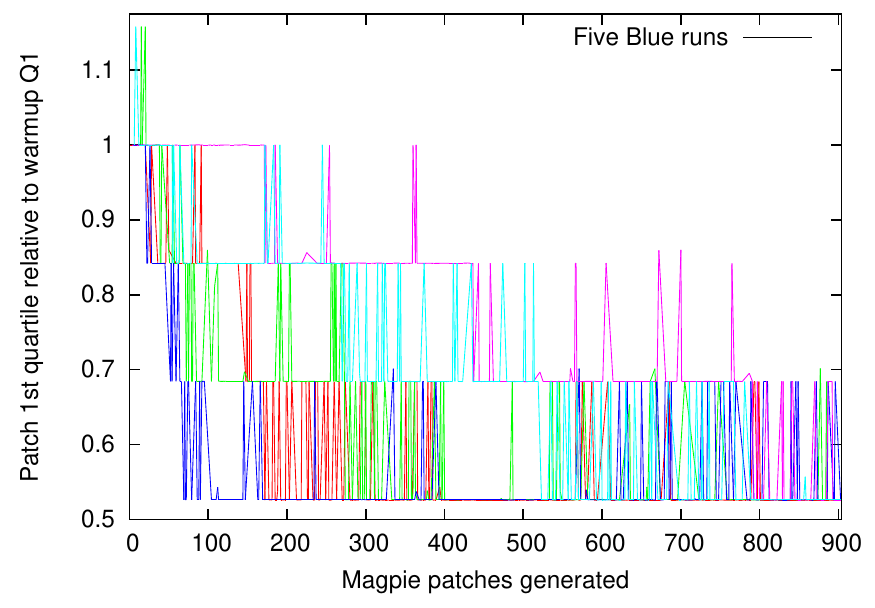}} 
\vspace*{-2ex}
\caption{
Blue patches. perf L1D counts for five Magpie runs relative to warm-up.
The fitness (lower is better)
of only patches which past all ten training cases is plotted.
\label{fig:blue_l1_fit}
}
\end{figure}

\begin{figure}
\centerline{\includegraphics{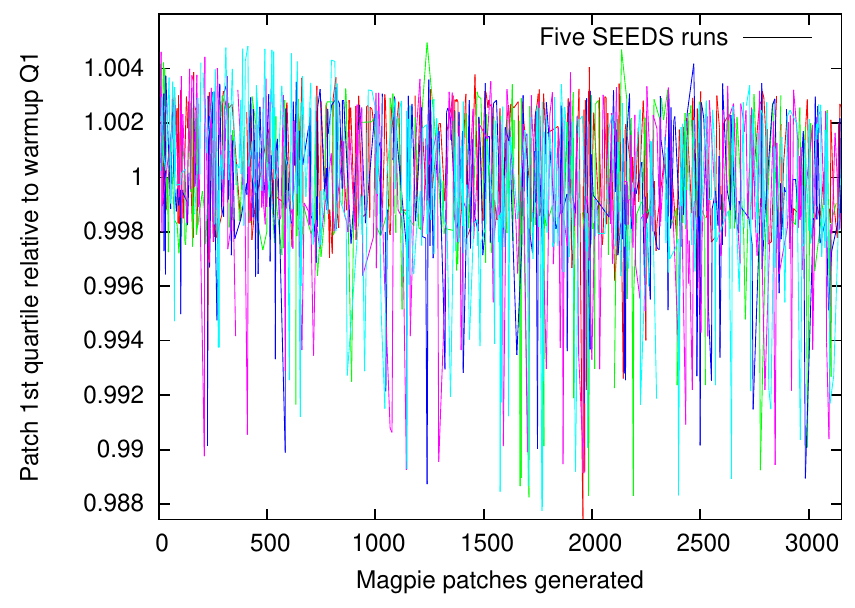}} 
\vspace*{-2ex}
\caption{
SEEDS patches. perf L1D counts for five Magpie runs relative to warm-up.
To suppress clutter data poor fitness, i.e.~above 1.005, are not plotted.
(Lower fitness, y-axis, is better.)
\label{fig:seeds_l1_fit}
}
\end{figure}

\begin{figure}
\centerline{\includegraphics{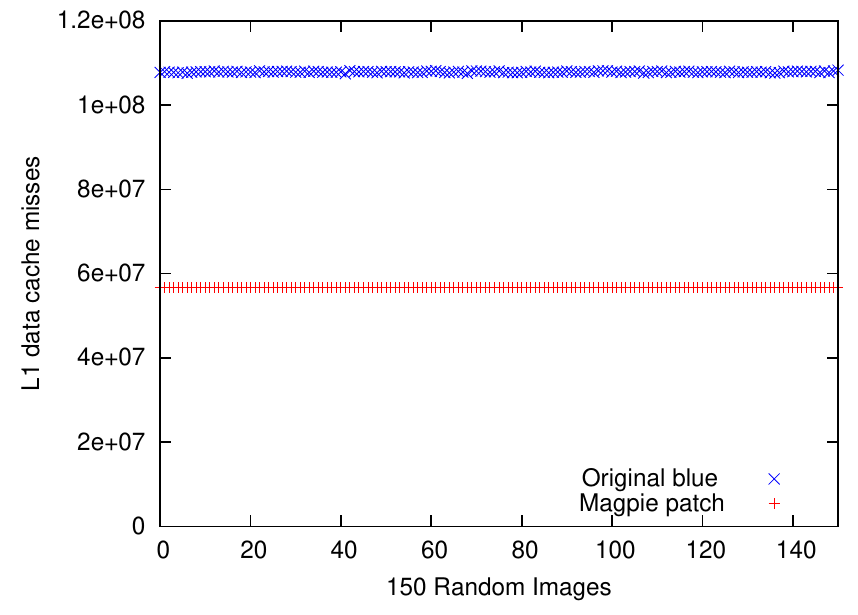}} 
\vspace*{-2ex}
\caption{
Magpie ``Blue'' patch out of sample performance
on 150 hold out images.
(Figure~\protect\ref{fig:blue_training} contains 
one of the training images.)
In all cases the patch returns the same count of blue pixels
as the original code,
but incurs only about half as many L1 data cache misses.
\label{fig:blue_l1_patch}
}
\end{figure}

\begin{figure}
\centerline{\includegraphics{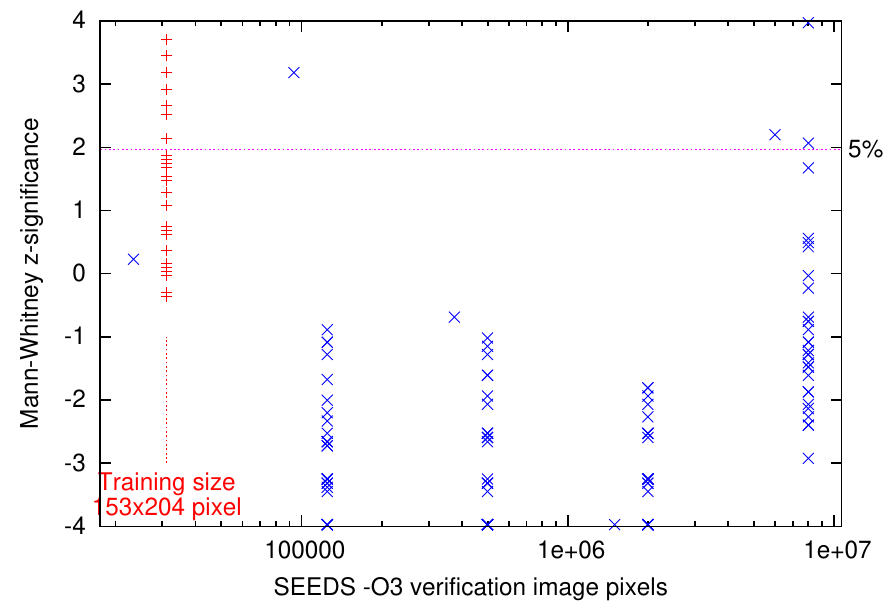}} 
\vspace*{-2ex}
\caption{
Magpie SEEDS patch out of sample performance
on 150 hold out images of various sizes
(e.g.~Figure~\protect\ref{fig:seeds_holdout}).
In all cases the patch segmented the image identically to OpenCV\@.
Measurement of L1 cache still very noisy even when repeated 11 times.
Nevertheless the patch tends to reduce L1 data misses on 
153$\times$204 images (red $+$) of the same size as the training image
(Figure~\protect\ref{fig:seeds_training})
and worse on larger images (blue~$\times$).
Note log x-axis.
\label{fig:seeds_l1_O3_patch}
}
\end{figure}

The results are summarised in Table~\ref{tab:cache_tab},
and 
Figures~\ref{fig:blue_l1_fit} and~\ref{fig:seeds_l1_fit}
show the fitness of Blue and SEEDS patches
found as Magpie was running. 
Note in particular that the results in columns~3 and~4 
(Mutant size and L1D)
are for the best fitness found by Magpie during its runs.
That is, in Table~\ref{tab:cache_tab},
the L1D improvement is as measured during training.
In the cases of the two geographic programs (OLC and H3),
while the patches retain their functional ability
to pass up to 10\,000 holdout tests,
the desired improvement in cache performance did not generalise.
Indeed it appears after taking care of the noise,
there is no real difference in L1 data cache misses
between the original and patch code.
This is in great contrast to the two data rich image 
programs,
where the patch 
does give reduced data cache misses
on images of the same size as the training data
(See Figures~\ref{fig:blue_l1_patch} and~\ref{fig:seeds_l1_O3_patch}).
Again both Blue and SEEDS patches also generalise in terms of still
giving the correct answer on unseen images.
However, the right-hand side of 
Figure~\ref{fig:seeds_l1_O3_patch}~(blue~$\times$)
shows the SEEDS patch does not give a reduction in L1 data cache misses
in images more than four times larger than the training image
(Figure~\ref{fig:seeds_training}).

On average 72\% of OLC patches fail to compile%
\footnote{Previously we used specialist mutation operators with \IR{}
which ensured all mutants compiled successfully to machine code
\cite{langdon:2023:EuroGP}.},
about 5\% fail one or more test cases,
while the remaining 23\% pass all ten fitness tests.
Table~\ref{tab:gather_l1_stats}
summarises the statistics for all four experiments.
The pattern for H3 and Blue is similar.
However the tabu list used with SEEDS
means whenever a patch fails a tabu check,
it is marked by Magpie as if it had failed to compile.
Hence the low figure for SEEDS' ok column%
\footnote{%
In \cite{langdon:2020:cec} we
used a similar idea to test if mutated
code is identical by inspecting the X86 assembler generated by
the GNU gcc compiler.
Also Mike Papadakis et al.~\cite{Papadakis:2015:ICSE}
compared compiler output to look for
equivalent mutants.}.
Note mostly patches which compile, run ok and pass all the tests.

\begin{table} 
\caption{
\label{tab:gather_l1_stats}
Summary of Magpie patches.
Average of five runs on each L1 data cache experiment.}
\vspace*{-2ex}
\centerline{%
\begin{tabular}{@{}lrrrrr@{}}
      & Compile error &Test failed  &time out     &too big      &ok          \\
\hline
olc   &        72\%&        5\%&        0\%&        0\%&       23\%\\
h3    &        61\%&        4\%&        0\%&        2\%&       33\%\\
blue  &        56\%&       13\%&        1\%&        0\%&       30\%\\
seeds &        87\%&       10\%&        0\%&        0\%&        2\%\\
\end{tabular}
}

\end{table}

Although Magpie has a nice tool for minifying patches,
we did not use it due to the noisy nature of our {\tt perf}
based fitness measure.

In all four experiments,
the final patch generated the same results as the original
program.
In the case of the two geographic tools (OLC and H3),
the holdout set contains ``missing data'',
i.e.\ postal addresses without a latitude, longitude location.
In these 85 cases OLC produces a default output,
while H3 aborts (with a non-zero Unix error code)
and an error message.
The H3 patches similarly detected and reported the error.
On the other 9915 holdout locations the patch similarly returns the
same output as the original H3.
That is both OLC and H3 patches pass all 10\,000 hold out tests.
However in neither case, were we able to show
the fitness seen in the Magpie runs,
translated to re-running them.
(See also Sections~\ref{sec:discuss_OLC}
and~\ref{sec:discuss_H3}.)

Our results are summarised in Table~\ref{tab:cache_tab}
column~5 ``L1D''
which gives the percentage reduction in L1 data cache misses
during training.
Unfortunately, as will be discussed in the next section,
the improvements in cache use reported during training with OLC and~H3,
did not generalise 
and out of sample, there is no reduction in L1 data cache misses.
In contrast on both image processing examples
we find significant (non-parametric Mann-Whitney test)
reduction in L1 data cache misses.
For the Blue bench mark it is 47\%.
And 1\% for the patch to the OpenCV image segmentation, SEEDS,
C++~code on 30 unseen 
unrelated images of the same size as the training image.

\pagebreak
\section{Discussion}
\label{sec:discuss}

\subsection{Types of Improvement Found}

\subsubsection{OLC}
\label{sec:discuss_OLC}

In one run Magpie found a single patch which on the ten
training locations
gave an 8\% reduction in L1 data cache misses.
It simply deletes one line.
The removed line is part of OLC's command line verification.
Since all the tests use well formed command lines,
checks for errors in parsing the command line are never triggered.
Therefor the deleted line is never used.
The 8\% improvement reported by Magpie 
appears to be just noise in the {\tt perf} measurements.
That is
(as with H3, next section),
the patch does not give sustained reductions in L1 data cache misses.
However,
as would be expected,
it does generalise functionally
and passes all 10\,000 hold out postal addresses
(shown with blue~$+$ on the left of Figure~\ref{fig:uk}).

\subsubsection{H3}
\label{sec:discuss_H3}

In one H3 run the best patch found contains 12 changes.
Two are {\tt Line\allowbreak{}Replacement},
nine are {\tt Line\allowbreak{}Deletion}
and there is one {\tt Line\allowbreak{}Insertion},
giving 14 line changes,
all of them non-functional.
Two delete lines of code that are never used.
One includes a header file, which due to conditional compilation has no effect.
Two add lines to {\tt const} arrays
which are never used.
Nine delete data in {\tt const} arrays,
in which only in two cases 
is the array referenced
and then in all cases, only parts of the array which are unchanged are used.
As would be expected,
the patched code passes all 10\,000 validation cases. 

\subsubsection{SEEDS}

In one run Magpie found a patch with seven changes
which together reduced reported L1 data cache misses by 7.4\%:

{\tt
Deletion~254, %
Deletion~204, %
Insertion~before~102~of~105,
Deletion~247, %
Insertion~before~87~of~66,
Deletion~63,  %
Insertion~before~137~of~109.}

In three places
GI exploited information available at run time but not to the GCC -O3 compiler
by using the fact that {\tt int~seeds\_prior} is always~2.
Thus line \verb'if( seeds_prior )' is always true and can be deleted
and the C++ \verb'switch( seeds_prior )' statement always takes \verb'case 2:'.
Therefore  \verb'case 1:' and \verb'case 5:' can be deleted
without changing the programs semantics.
In fact so too can 
the \verb'case 3:' and \verb'case 4:' statements.
Indeed in
other runs Magpie removed the \verb'case 3:' and/or \verb'case 4:'.
(These redundant \verb'case' statements are on
lines~247--249 and~254.)

These can be seen as traditional GI speed-up changes,
in which unnecessary operations are removed.
They may also have a potentially beneficial impact on the data cache,
although only removing \verb'if( seeds_prior )' on line~63 has a direct 
impact on data usage.
Reducing the code size may have a subtle second order impact
on the data cache
by changing the machine code generated by the optimising compiler.

Deletion of line~204 does the opposite,
since it removes \linebreak[4]
\verb'if( labelA != labelB )'
which is only true infrequently.
Thus the patch causes {\tt update()} on the next line
to be called more often.
But each time it calculates the same label for pixel~(w-1,y)
and so does not change the image's segmentation.
In the fitness function there is no trade-off:
the patch can waste as many CPU instructions as it likes,
only reducing L1 data cache misses are important.
The {\tt update()} following line~204
is called in a loop of 153 iterations
in which data adjacent to (w-1,y),
i.e.~(0,y+1),
is needed on the next loop iteration.
Therefore
the unnecessary call of {\tt update()} may
have the beneficial impact of keeping data in the L1 cache%
\footnote{Cache lines are 64 bytes and so can hold 
labels for 16 horizontally adjacent pixels.}.
This argument aught to hold with images of any size,
but Figure~\ref{fig:seeds_training} suggests the
overall benefit to the L1 data cache does not hold each time the image
size is increased by a factor of four.
This may be because with larger images 
the other arrays {\tt update()} uses,
nullify the benefit of
trying to keep the label pixel data in the L1 cache.

The other three changes insert copies of lines of code
which again do not change values but cause them to be recalculated:

\verb'priorB = threebyfour(x, y, labelB);' \hfill(before line~87)

\verb'int a11 = Labels((y - 1) * width + (x - 1));' \hfill(line~102)

\verb'int a23 = Labels((y) * width + (x + 1));' \hfill(line~137)

\noindent
The compiler recognises that 
\verb'int a11' and \verb'int a23' are unused,
however this does not ensure it ignores them and so
it may generate code to access labels for pixels (x-1,y-1)
and (x+1,y).
Both occur in a doubly nested loop which scans the image
in the wrong order:
i.e.~the inner loop samples the data a long way apart,
rather than accessing neighbouring cache lines.
Hence the available 512 L1 cache lines may be under great stress.
For the patch adding \verb'int a11', it seems label (x-1,y-1) is often
needed almost immediately,
so duplicating the calculation of \verb'int a11' perhaps costs little in
terms of cache usage.
Whereas the \verb'int a23' patch reads the label for pixel (x+1,y)
which has been recently read, and so again
recalculating \verb'int a23' may impose little on the cache.
The nested loops makes analysis hard and there may be
important data locality effects,
similar those suggested for {\tt update()} in the earlier paragraphs.
These effects may totally overwhelm the computer's 512 L1 data cache lines
in bigger images.

In contrast the additional call to
\verb'threebyfour(x, y, labelB);'
to set \verb'priorB'
inserted before line~87,
is in a pair of nested loops which accesses pixel data in the ``right'' order,
i.e~along the \mbox{x-axis}.
If called,
\verb'threebyfour' reads four adjacent pixel labels in three adjacent rows.
These will occupy between three and six cache lines
(depending how the x-dimension straddles 64~byte boundaries).
Like the line~204 patch discussed above,
which causes {\tt update()} to be called many more times,
given the nested loops and that \verb'threebyfour' has been recently called,
the extra call to \verb'threebyfour'
may be beneficial in keeping data
(which is already in the L1 cache) in it.
Data near these $3\times4$ pixels,
and hence in the same cache lines,
is almost certain to be needed immediately 
by the next loop iteration.

\subsection{Non-stationary noise,
Profiling, Mutation and other Search Operators}
\label{sec:discuss2}

\begin{figure}
\centerline{\includegraphics{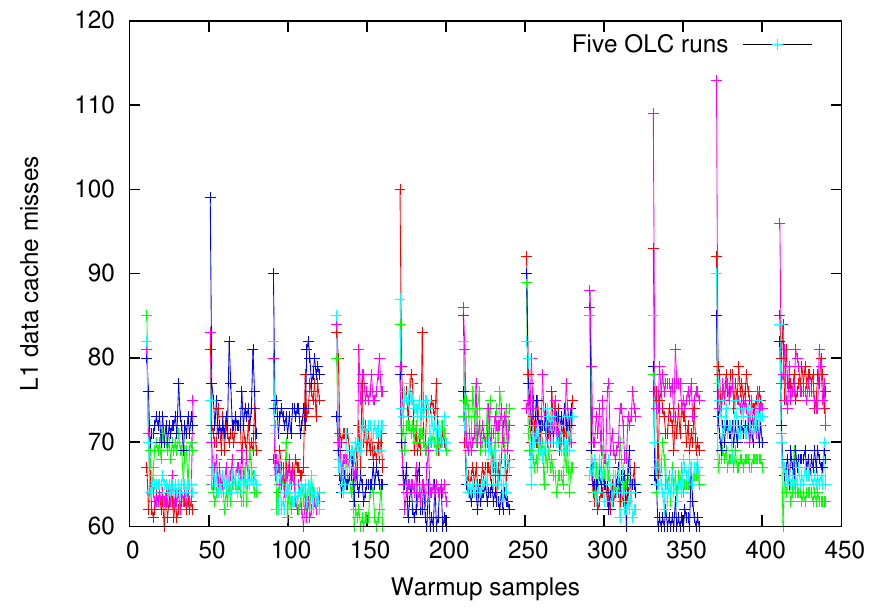}} 
\vspace*{-2ex}
\caption{
L1 data cache misses for Magpie's OLC warm-up
(when the same program is repeatedly measured).
Notice:
the presence of outliers,
that often the first measurement of a group is much bigger than the rest,
and the systematic, as well as random, variation between runs
of the same software on the same computer.
\label{fig:olc_l1_warmup}
}
\end{figure}

As mentioned in Section~\ref{p.Q1},
motivated by large positive outliers
often seen in run time measurement,
we have used the first quartile in the fitness function,
as it is a robust statistic which is 
relatively immune to positive outliers.
Although Figures~\ref{fig:validate_olc_l1}
and~\ref{fig:validate_h3_l1}
do not plot the small number of large positive outliers,
they show that the distribution of cache misses is very noisy
and  also that it is more symmetric and Gaussian like than expected.
Given the apparent symmetry 
it may be that the median would give a more
consistent fitness measure than the first quartile
(see also Figure~\ref{fig:olc_l1_warmup}).

However, the multi-modal distributions shown in
Figures~\ref{fig:validate_olc_l1} and~\ref{fig:validate_h3_l1}
with different colours,
show another little discussed problem:
The L1D noise is {\em not} stationary~\cite{Moskowitz:thesis},
but subject to some unknown drift.
Classical arguments,
which assume multiple measurements are
independent and identically distributed (IID),
suggest increasing the number of measurements $n$
will reduce the impact of noise in proportion to~$\sqrt{n}$.
However this misses the fact if the noise is non-stationary,
then measurements taken during a Magpie run
(or indeed during any evolutionary computing~EC run)
will drift.
Not only during the EC run itself,
but also when the evolved artifact is used.
It may be we need,
not only to take multiple L1D measurements per fitness evaluation,
but also, during the EC run, to make estimates of the drift.
Perhaps this might be done by running some known fixed example code.
If online drift estimation turns out to be effective,
it is likely that effort spent on combating noise during
EC runs would be well worth while,
as it should lead to better more robust solutions.

Magpie has targeted whole functions that could possibly be called.
Particularly in our largest example, H3,
this is not sufficient.
Since there are both lines of code and pre-set data values
that are never used,
but Magpie is wasting effort on trying to optimise them.
It is common in GI
to profile the program to be evolved,
and then to target only code that is indeed executed
\cite{Langdon:2013:ieeeTEC}.
In these examples, profiling was not used.
Indeed the presence of a huge volume of unused data in H3
hints at a further problem.
Earlier profiling has concentrated upon code execution.
It appears not to have considered that data might be created 
(and so need maintaining)
which is not used by the code 
during every-day mundane operations.

Of course Magpie has more sophisticated syntax aware approaches 
and they also might benefit from profile data.
Apparently more recent version of Magpie,
already rule out simple patches
that move \#include files
or simply swap lines containing just a single curly bracket~``\verb/}/''.
These may help,
but we fear that as usual, fitness driven evolution
will find a way to exploit code changes
by making other, as yet unthought of, 
apparently ``useless'' changes.

The trick of enforcing that each patch make a new semantic
change by keeping a tabu list
(as was used with SEEDS)
will not deal with the problem of
wasted effort being spent on generating patches
that mutate either unused code or unused data.
The tabu list it uses to prevent duplicates
is based on the object file created by the compiler.
Changes to either unused code or unused pre-set data
would change the object file and so although useless would
appear to be plausible semantic changes.
The problem is exacerbated here with Magpie's local search,
as apparently ``useless'' changes to \mbox{un-executed} code or
unused data
can, due to the noisy fitness function, appear to be beneficial
and so drive search in unproductive directions.
However we need to be cautious,
as the behaviour of caches is often proprietary
and the exact implications of even ``obviously useless''
changes is in practice unknowable.
For example,
even when the GCC compiler (with -O3) issues a warning 
saying the patch introduces an
``unused variable'',
it may change the machine code it generates.
So potentially changing the behaviour of the caches.

So far with Magpie we only used a few types of mutation:
{\tt Line\allowbreak{}Deletion}%
\footnote{%
Delete is commonest way programmers speed up code~\cite{callan2022}.},
{\tt Line\allowbreak{}Insertion} and {\tt Line\allowbreak{}Replacement}.
Many others are feasible.
Similarly we have only used Magpie's local search,
and other strategies could be explored.
Indeed Magpie already supports genetic programming.
In future Magpie's parameter search might also be applied to
aspects outside the source code,
such as the compilation and linking.

\section{Conclusions}
\label{sec:conclude}

We have taken a new open source genetic improvement tool
written in Python
(Magpie)
and applied it to industrial C source codes
from Google and Uber,
the ``Blue'' image processing problem
and OpenCV's state-of-the-art image segmentation C++ code.
In particular to the never before attempted
optimisation of the deepest level of the hardware data cache hierarchy,
which is essential to modern computers. 
For OpenCV's SEEDS
a 1\% reduction
on compiler -O3 optimised code
was found,
whilst ``Blue'' L1 data cache misses were almost halved
(47\%).

In retrospect perhaps it was optimistic to
expect search to find ways to make a large impact on
the data cache on our 
two industrial geo-positioning examples
(OLC and H3).
After all they have a large amount of code for
only a very small data input (two floats).
Perhaps more importantly a typical run only incurs 
a few hundred L1 data cache misses,
making it a small target to improve.
Nonetheless it is heartening that 
in a minute or a few hours%
\footnote{%
Magpie with the GNU C/C++ compiler (-O3) processed
between 0.7 and 10 patches per second
depending on OLC, H3, Blue or SEEDS\@.
Whereas
with Clang 14.0.0 in our earlier work~\cite{langdon:2023:EuroGP}
on OLC and H3,
we processed between 0.25 and 1.5 \IR{} patches per second,
depending on experiment and if using -O3 or not.}
on compiler optimised code
improvements were found and that these at least generalise functionally.

\begin{acks}
I am grateful 
to 
H.Wierstorf 
for help with
\href{http://www.gnuplotting.org/plotting-the-world-revisited/}
{gnuplot}. 

\noindent
Supported by 
\href{https://gtr.ukri.org/projects?ref=EP%2FP023991%2F1}
{EP/P023991/1}
and the Meta OOPS project.
\end{acks}

\bibliographystyle{ACM-Reference-Format}

\bibliography{cache}

\end{document}